\documentclass[a4paper,UKenglish,cleveref, autoref, thm-restate]{lipics-v2021}

\hideLIPIcs  

\usepackage{times}
\usepackage{soul}
\usepackage{url}
\usepackage{amsmath}
\usepackage{amssymb}
\usepackage{amsthm}
\usepackage{mathtools}
\usepackage{booktabs}
\usepackage{cleveref}
\usepackage{bbm}
\usepackage{multirow}
\usepackage{stmaryrd}
\usepackage{numprint}
\usepackage{enumitem}
\usepackage{caption}
\usepackage{subcaption}
\usepackage{algorithm}
\usepackage[noend]{algpseudocode}

\usepackage{tikz}

\newcommand{\hide}[1]{}

\usepackage{todonotes}

\title{Holy Grail 2.0: From Natural Language to Constraint Models}

\author{Dimos Tsouros}{KU Leuven, Belgium}{dimos.tsouros@kuleuven.be}{0000-0002-3040-0959}{}
\author{Hélène Verhaeghe}{KU Leuven, Belgium}{helene.verhaeghe@kuleuven.be}{0000-0003-0233-4656}{}
\author{Serdar  Kad{\i}o\u{g}lu}{AI Center of Excellence, Fidelity Investments, USA\\
Department of Computer Science, Brown University, USA}{serdar.kadioglu@fmr.com}{0000-0002-4672-6830}{}
\author{Tias Guns}{KU Leuven, Belgium}{tias.guns@kuleuven.be}{0000-0002-2156-2155}{}

\authorrunning{D. Tsouros, H. Verhaeghe, S.  Kad{\i}o\u{g}lu and T. Guns} 

\Copyright{Dimos Tsouros, Hélène Verhaeghe, Serdar  Kad{\i}o\u{g}lu and Tias Guns} 



\category{} 

\relatedversion{} 

\funding{This research received funding from the
European Research Council (ERC) under the European
Union’s Horizon 2020 research and innovation program
(Grant No. 101002802, CHAT-Opt) }

\nolinenumbers 

\EventEditors{John Q. Open and Joan R. Access}
\EventNoEds{2}
\EventLongTitle{42nd Conference on Very Important Topics (CVIT 2016)}
\EventShortTitle{CVIT 2016}
\EventAcronym{CVIT}
\EventYear{2016}
\EventDate{December 24--27, 2016}
\EventLocation{Little Whinging, United Kingdom}
\EventLogo{}
\SeriesVolume{42}
\ArticleNo{23}

\begin{document}


\maketitle

\begin{abstract}
Twenty-seven years ago, E. Freuder highlighted that "Constraint programming represents one of the closest approaches computer science has yet made to the Holy Grail of programming: the user states the problem, the computer solves it".
Nowadays, CP users have great modeling tools available (like Minizinc and CPMpy), allowing them to formulate the problem and then let a solver do the rest of the job, getting closer to the stated goal. 
However, this still requires the CP user to know the formalism and respect it.
Another significant challenge lies in the expertise required to effectively model combinatorial problems. 
All this limits the wider adoption of CP. 
In this position paper, we investigate a possible approach to leverage pre-trained Large Language Models to extract models from textual problem descriptions. 
More specifically, we take inspiration from the Natural Language Processing for Optimization (NL4OPT) challenge and present early results with a decomposition-based prompting approach to GPT Models.
\end{abstract}

\section{Introduction}

In 1996, Eugene Freuder highlighted that "\textit{Constraint programming represents one of the closest approaches computer science has yet made to the Holy Grail of programming: the user states the problem, the computer solves it.}"~\cite{freuder1996pursuit}.
Twenty-seven years later, current technologies like modeling languages (such as Minizinc~\cite{nethercote2007minizinc} or Essence~\cite{frisch2008ssence}) and modeling libraries (like CPMpy~\cite{guns2019increasing}) have proven that indeed, constraint programming has the capability to achieve this holy grail of programming. 

However, there is still a gap between a natural formulation of the problem (in Natural Language) and a CP model.
The user is still responsible for transforming the problem at hand as an optimization model, to model it as a set of variables with their domains, a set of constraints, and an objective function. 
This is not always trivial and requires quite some expertise, and it is considered a bottleneck for the wider adoption of CP.



Today, with the astonishing recent advances in Large Language Models (LLMs), isn't it the time to think ahead, toward the holy grail 2.0, where the user could use natural language to define a problem, and the system would be able to understand and automatically extract the formal model required by the solvers?


Inspired by the developments in the area of NL4OPT, and especially the recent advancements in the use of LLMs to process natural language, recent works have started investigating methods for automating the "natural language to optimization model" process using LLMs, to be more efficient and accessible to non-experts. The whole process of modeling a problem is usually split into different subtasks, as it is a multi-step process, and LLMs usually find it hard to tackle such tasks.

In~\cite{dakle2023ner4opt}, the NER4OPT subtask is formalized, as an interdisciplinary problem at the intersection of Named Entity Recognition (NER) and Combinatorial Optimization. The differences from standard NER tasks are discussed, and a method to tackle this subtask is proposed, using classical techniques
based on morphological and grammatical properties combined with modern methods leveraging pre-trained LLMs.
A system that tries to close the modeling loop and automatically formulate optimization models from problem descriptions is proposed in ~\cite{ramamonjison2022augmenting}. This system uses a two-step approach. First, the problem description is transformed into an intermediate representation, which is very similar to the NER4OPT task, i.e., labeling the entities of the problem. Then, in the second step, this intermediate representation is used to formally formulate the optimization problem. This system is built upon the BART language model~\cite{lewis2019bart}.

In 2022, the NL4Opt competition \cite{ramamonjison2023nl4opt} took place at the NeurIPS conference. 
They targeted the two tasks described above, i.e., recognizing the semantic entities corresponding to the optimization problem (i.e., variables, values, objective) and generating a meaningful representation (i.e., identifying the actual relations between variables, the associations of the domains and variables,...). For tackling the second subtask, the system from~\cite{gangwar2022highlighting} managed to achieve the best results by splitting it again into subtasks, i.e. first find the relations between the entities of the problem, and only then continue with the \textit{formulation} task, as it was shown that LLMs can exploit this information to improve their performance. Again, BART was used in this work.

The target formulation for the mentioned works focused mainly on creating linear programming problems. A first approach for modeling constraint problems using CP, involving also the translation to the Minizinc modeling language~\cite{nethercote2007minizinc}, was presented in~\cite{almonacid2023towards}. The described system uses a single-step approach for the modeling, followed by an automated fixing process, that compiles and debugs the generated model in the Minizinc language. The early results are promising, however, in some cases the specifications of the optimization problem are not achieved, due to the one-step modeling method.  

Inspired by the developments in the area of NL4OPT, the recent advancements in the use of LLMs (such as GPT-3.5), and the recent works that show promising results, in this position paper we propose a modular step-by-step framework to model a problem based on the text description. In this framework, we combine the benefits of the mentioned systems, by splitting the modeling task into four subtasks that have been shown to boost the performance of LLMs. The system takes as input the description in natural language, and then the modeling process starts. First extract the problem's entities (variables, domains, constraints, objective), then find the relations between the entities, use the entities and their relations to formalize it as an optimization problem, and translate it into a constraint model language (such as CPMpy). 
After the modeling is finished, the resulting code is compiled and run, and automatically debugged if necessary. Finally, the system will interact with the user to refine the model. 

The rest of the paper is structured as follows:
In Section~\ref{sec:framework} we discuss our proposed method for automatic modeling by exploiting the emerging abilities of LLMs and we describe the different subtasks. Section~\ref{sec:lever} focuses on how LLMs can be leveraged and discusses techniques to boost them through prompt engineering. Then, we discuss the different levels of abstraction of problem descriptions that we want to tackle (Section~\ref{sec:abstraction}). In Section~\ref{sec:ex} we give an example of how our system works. Section~\ref{sec:concl} summarizes our paper and gives some directions for future work.







\section{Exploiting LLMs for converting natural language to constraint models}
\label{sec:framework}


Figure~\ref{fig:diagram} shows the high-level loop for modeling constraint problems from text descriptions. The modeling task can of course be done altogether using LLMs, as in~\cite{almonacid2023towards}. However, the modeling task is not a trivial task and consists of many subtasks, as discussed in~\cite{ning2023novel,dakle2023ner4opt}, and LLMs find it difficult to handle difficult multi-step tasks in one step~\cite{rae2021scaling}. This is confirmed by early experiments with the one-step approach.
We hence propose a decomposition-based approach, inspired by recent work in NL4OPT. In our proposed approach the system splits the modeling task into the four following steps:

\begin{figure}[tb]
	\begin{center}
		\includegraphics[width=\textwidth]{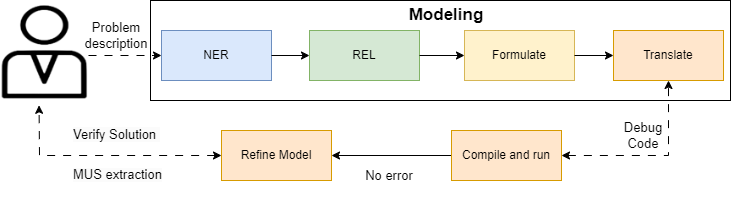}
	\end{center}
	
	\caption{The framework proposed} \label{fig:diagram}
\end{figure}

\begin{enumerate}

    \item 
\textbf{NER4OPT}
The modeling system has to extract the semantic entities of the optimization problem, i.e. the parameters, variables, domains, constraints, and objective. 
Note that this task differs significantly from the classical NER in NLP due to its multi-sentence dependency with high-level of ambiguity, low data regime with high-cost of annotation, and inherent aleatoric uncertainty. It is coined as Ner4Opt, an interdisciplinary problem at the intersection of NLP and Optimization and studied in detail for its lexical, semantic, and hybrid solutions~\cite{dakle2023ner4opt} using large-language models as well as fine-tuning with optimization corpora. 

\item 
\textbf{REL}
Note that, in NER4OPT the goal is to label words in the description using predefined optimization entities. However, the relations between the entities are not yet found, e.g. which variables are in the scope of each constraint.  
In the second task, the system has to identify the relations between the various entities extracted in the previous step, link the values with the type of parameters, find the domains associated with the variables, and identify the scopes of the constraints. This subtask is similar to the first step from~\cite{gangwar2022highlighting}, which takes as input the labeled entities and splits the process of formulating the optimization problem into two steps.

\item 
\textbf{Formulation}
The third subtask of our framework is to formulate the problem as a constraint problem, given the list of labeled entities and their relation, in a formal way. This subtask is again inspired by~\cite{gangwar2022highlighting}, as it follows their step of finding the relations of the constraints.

\item 
\textbf{Translation}
Then, fourth, the system would take as input the problem formulation from the previous step and translate it into code, following the syntax of the desired modeling language, as in~\cite{almonacid2023towards}. Given a clear description of the coding task, with the required entities, constraints etc., LLMs have been shown to be effective in code-writing~\cite{xu2022systematic}.

\end{enumerate}

After the modeling process is finished, with the output being the code in any modeling language desired, we still cannot be sure that the extracted model is correct. First of all, the code given could contain errors either in compilation or in runtime. In addition, the model could have missed some constraints or modeled some wrongly. Finally, the user may have given an unsatisfiable problem and wants to refine it. Hence, we propose two additional steps to close the loop.

\begin{enumerate}
    
\item 
\textbf{Fixing the output}
The fifth step would be compiling (and running) the code, and identifying any errors. In this case, the system automatically corrects the errors in a loop, until a bug-free code is extracted. This step is inspired by the method presented in~\cite{almonacid2023towards}. It has been shown that LLMs can be exploited for bug-fixing with good results~
\cite{sobania2023analysis}.

\item 
\textbf{Refining the model}
Finally, the last step would be the presentation of the final model and potential solution(s) to the user. 
The user would then be able to verify the solution or request Minimal Unsatisfiable Subset(s) (MUS) and/or explanations in case there is no solution. 
Interactions with the user at that step would be done in order to provide corrections or precisions to the model in order for the user to obtain the desired solution.
\end{enumerate}



With such a modular process, our goal is also to allow people to adapt our process to their needs. 
One of the use cases is for the 4th step. 
We plan on targeting the CPMpy modeling language to execute the transformation from the formal model info code. 
However, another user could want the system to output Minizinc code.
In that case, they would have just to replace this module.
Also, we should be able to change the LLM used in an easy way, or use hybrid methods, e.g. using the method from ~\cite{dakle2023ner4opt} in the first step.

\section{Leveraging LLMs}
\label{sec:lever}


Our framework will be highly using LLMs, being also capable to use specialized tools to tackle some subtasks, like for the NER4OPT task~\cite{dakle2023ner4opt}
Multiple LLMs will be tested such as GPT variations or LLAMA. 
Prompt engineering (also known as in-context learning), which is the task of carefully designing prompts to better leverage LLMs capabilities, has been shown to significantly boost performance on several tasks~\cite{reynolds2021prompt}. There are many different ways to use prompt engineering for a task on hand. We mainly experimented and included in our system the following: 

\begin{itemize}

\item 
\textbf{Roles and goals}~\cite{schmidtcataloging}. One efficient way to get better results using prompt engineering is to specify the role the LLM has to play and the goal to achieve.
The intent of this technique is to localize the training of the LLM for the specific task on hand, selecting what
types of output to generate and what details to focus on.
Some LLMs have also specific internal tools that can help to achieve that in a better way. For example, for the GPT models, we can make use of the system/user prompt duality. 
The system prompt allows us to set up the role and goals, i.e., we can give a statement such as "Assume you are a combinatorial optimization expert and you need to model a combinatorial problem as an optimization problem". Then the user prompts are used to converse with it. 

\item 
\textbf{Few-shot learning}~\cite{brown2020language}  will be very useful to boost the performance of LLMs, by giving examples of the task on hand in each step and how to solve it. In addition, it can be used for specifying the formatting we require for the output when formulating the formal model, for example. For instance, the extractions from pre-trained Ner4Opt models can be used to automate the few-shot example generation for in-context learning. 

\item 
\textbf{Chain-of-thought}~\cite{wei2022chain} prompting techniques take prompting one step further,
allowing models to decompose multi-step problems into intermediate steps. This means that additional computation can be allocated to problems that require more reasoning steps than simple input-process-output scenarios. It has been shown that even zero-shot Chain-of-thought prompts can be very efficient, especially in symbolic tasks.

\item 
\textbf{Tree of Thoughts}~\cite{long2023large, yao2023tree} techniques are inspired by the human mind's approach to solving complex reasoning tasks through trial and error. In this process, the system explores the solution space through a tree-like process, using backtracking when needed. This helps to generate alternative outputs for a given prompt, choosing the best one, in order to ensure better results.

\item 
\textbf{Plan-and-Solve}~\cite{wang2023plan}.
Despite the remarkable success of Zero-shot Chain of Thoughts in solving multi-step reasoning tasks, there are still many cases where
Intermediate reasoning step(s) are missed, leading to worse performance. This happens especially when there are many steps involved in the task on hand. Plan-and-Solve was proposed to alleviate this issue. This method consists of 2 different steps:  First, the system is asked to focus on devising a plan dividing the entire task, without the need to solve the problem.
Then, in a separate step, it is asked to carry out the subtasks according to the plan. 

\end{itemize}

Note that, as the ways these techniques are used are not task-specific, they can be included in a system that automatically transforms the prompt given by the user to a new prompt using the above techniques. So, the real user of such a system does not have to focus on prompt engineering and just on the task on hand, in our case on modeling the problem.







\section{Levels of abstraction}
\label{sec:abstraction}

In addition to using existing datasets, such as the one from the NL4Opt competition \cite{ramamonjison2023nl4opt}, one of our goals is also to measure how well the framework is able to model and recognize problems. 
To evaluate this, our plan is to evaluate the system in various levels of abstraction.

\begin{enumerate}

    \item The first level of abstraction would be the easiest, where the problem name would be clearly stated (i.e. use of the words tsp, knapsack, graph coloring,...), and the constraints and variables already identified using tokens. 
    This is the baseline. 
    \item In the second level, we would omit the name of the problem while still explicitly describing the variables, constraints, and parameters of the problem.
    This level will help us see if problems are recognized.
    \item In the third level, the known lexical of modelization (i.e. words like constraint, variables, domains,...) would be removed. 
    Here, our wish is to measure how well the components of the model can be without being pre-identified using tokens.
    \item The fourth level would be the more abstract one. Some parameters could be implicit and not numerical for example. This would be the closest to what any human unaware of what an optimization problem is would do.
    This final level allows us to really see whether LLMs can do something when the structure of the problem disappears totally.
\end{enumerate}

It is to be expected that the more abstract we are in the problem descriptions, the more difficult it would be for the LLMs and the more interactions it would need with the user to get the perfect model and solution. Simple examples of the different levels of abstraction are given in Example~\ref{ex:prompts}.

\begin{example}
\label{ex:prompts}
Problem descriptions with different levels of abstraction for the same model:
\begin{itemize}
    \item [(L1)] \textit{I wish to solve a Knapsack problem, where I have 5 items, and so 5 binary variables to tell me which item is in or not. The weights of my items are 2, 3, 7, 4, and 1. The utilities of my items are 2, 3, 1, 2, and 3. The limit of weight is 10. Can you model it?}
    \item [(L2)] \textit{I wish to solve a problem, where I have 5 items, and so 5 binary variables to tell me which item is in or not. The weights of my items are 2, 3, 7, 4, and 1. The utilities of my items are 2, 3, 1, 2, and 3. I wish to limit the total weight to 10. Also, I wish to maximize the utility of the objects I take. Can you model it?}
    \item [(L3)] \textit{I wish to go on vacation. The airport only allows 10 kg for my suitcase. I have 5 items. The weights of my items are 2, 3, 7, 4, and 1. The utilities of my items are 2, 3, 1, 2, and 3. Can you tell me what items I should take in order to pass the best vacation?}
    \item [(L4)] \textit{I wish to go on vacation. The airport only allows 10 kg for my suitcase. I have 5 items: my ski combination, weighing 7 kg, some warm clothes, weighing 4 kg, some hiking boots, weighing 3 kg, a book on hiking, of 1 kg, and some umbrella, of 2 kg. As I'm going hiking, I think my boots and my book are really important, while the ski combination would not help me well. Can you tell me what items I should take in order to pass the best vacation?}
\end{itemize}
\end{example}


\section{Usage Example}
\label{sec:ex}

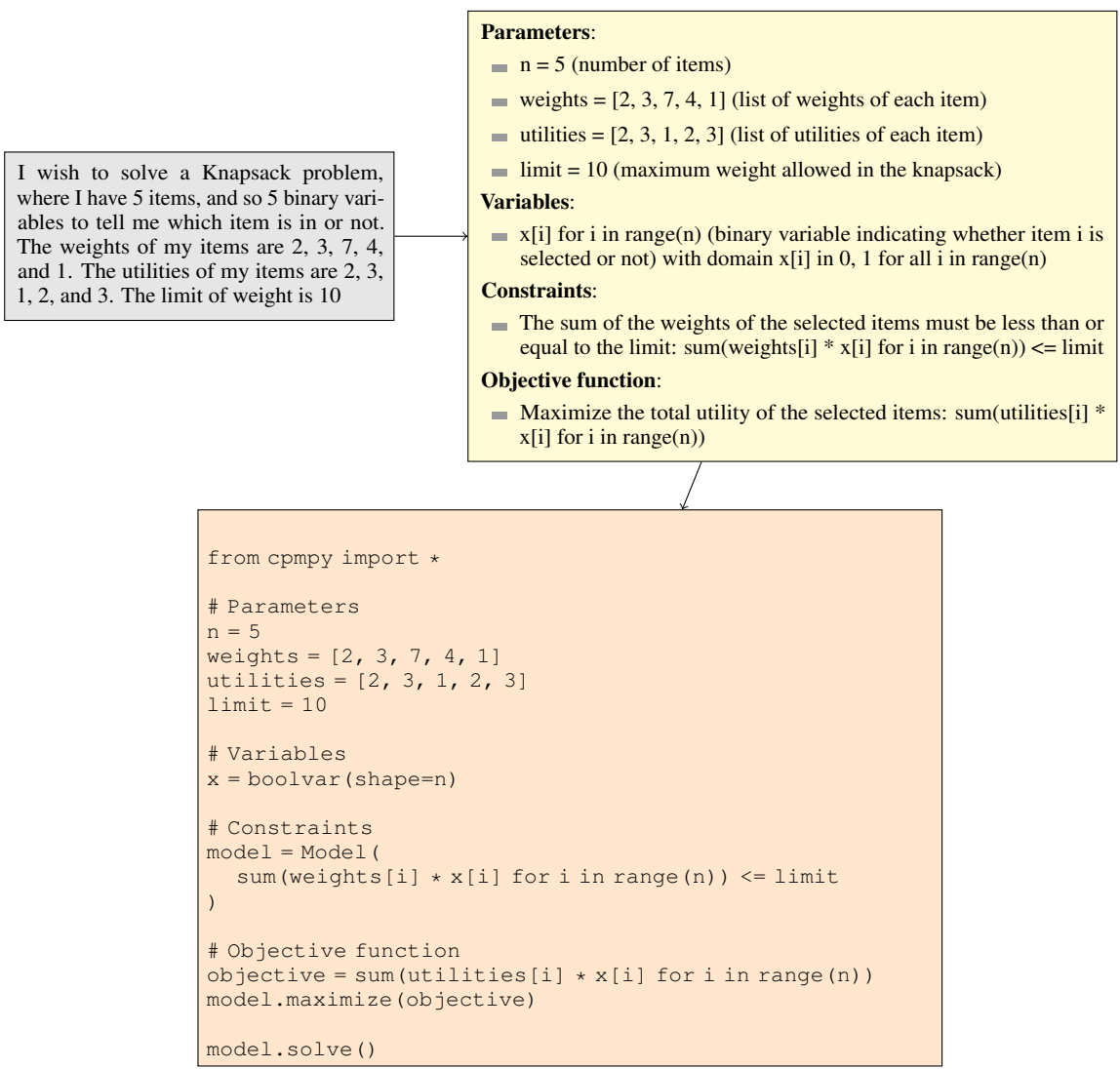
\begin{figure}
  \centering
  \begin{tikzpicture}

  \node[inner sep=5pt, fill=gray!20, draw=black, font=\footnotesize] (description) at (-10,0) {
      \begin{minipage}{0.35\textwidth}

I wish to solve a Knapsack problem, where I have 5 items, and so 5 binary variables to tell me
which item is in or not. The weights of my items are 2, 3, 7, 4, and 1. The utilities of my items are
2, 3, 1, 2, and 3. The limit of weight is 10

      \end{minipage}
        };      
        
    \node[inner sep=5pt, fill=yellow!20, draw=black, font=\footnotesize] (formulation) at (-2,0) {
      \begin{minipage}{0.6\textwidth}

\textbf{Parameters}: 
\begin{itemize}

\item n = 5 (number of items)
\item weights = [2, 3, 7, 4, 1] (list of weights of each item)
\item utilities = [2, 3, 1, 2, 3] (list of utilities of each item)
\item limit = 10 (maximum weight allowed in the knapsack)
\end{itemize}

\textbf{Variables}:
\begin{itemize}
\item x[i] for i in range(n) (binary variable indicating whether item i is selected or not) with domain x[i] in {0, 1} for all i in range(n)
\end{itemize}

\textbf{Constraints}:
\begin{itemize}
\item The sum of the weights of the selected items must be less than or equal to the limit: 
    sum(weights[i] * x[i] for i in range(n)) <= limit
\end{itemize}

\textbf{Objective function}:
\begin{itemize}
\item Maximize the total utility of the selected items:
    sum(utilities[i] * x[i] for i in range(n))
\end{itemize}

      \end{minipage}
    };

    \node[fill=orange!20, draw, text width=0.7\textwidth, font=\footnotesize] (codebox) at (-5,-7.5) {
      \begin{verbatim}from cpmpy import *

# Parameters
n = 5
weights = [2, 3, 7, 4, 1]
utilities = [2, 3, 1, 2, 3]
limit = 10

# Variables
x = boolvar(shape=n)

# Constraints
model = Model(
    sum(weights[i] * x[i] for i in range(n)) <= limit
)

# Objective function
objective = sum(utilities[i] * x[i] for i in range(n))
model.maximize(objective)

model.solve()      \end{verbatim}
    };    
    
    \draw[->] (description) -- (formulation);
    \draw[->] (formulation) -- (codebox);

  \end{tikzpicture}
  \caption{Example with the 1st level of abstraction}
  \label{fig:ex1}
\end{figure}

Here we provide examples of the input-output of our system. The system utilized here uses GPT-3.5, with prompt engineering for all different subtasks. We use CPMpy as the modeling language to translate in the 4th step. We will use as examples the descriptions used in Example~\ref{ex:prompts} for the first (Figure~\ref{fig:ex1}) and fourth (Figure~\ref{fig:ex2}) levels of abstraction, so we can examine the resulting models in the two extreme cases.  

As we can see, in both cases the resulting model is correct, and the code provided is working. In the first case, where the description of the problem directly describes the problem as a knapsack problem, giving the specific weights and utilities, it was easier to extract the respective parameters of the model. However, in the second case, the problem type is not directly given, and neither are the utility values. The system is still able to extract the entities of the problem correctly, connect the variables with the respective weights, build the knapsack constraint correctly, and, importantly, infer utility values to build an objective function.

\begin{figure}
  \centering
  \begin{tikzpicture}

  \node[inner sep=5pt, fill=gray!20, draw=black, font=\footnotesize] (description) at (-10,0) {
      \begin{minipage}{0.35\textwidth}

I wish to go on vacation. The airport only allows 10 kg for my suitcase. I have 5 items: my ski combination, weighing 7 kg, some warm clothes, weighing 4 kg, some hiking boots, weighing 3 kg, a book on hiking, of 1 kg, and some umbrella, of 2 kg. As I'm going hiking, I think my boots and my book are really important, while the ski combination would not help me well. Can you tell me what items I should take in order to pass the best vacation?

      \end{minipage}
        };      
        
    \node[inner sep=5pt, fill=yellow!20, draw=black, font=\footnotesize] (formulation) at (-2,0) {
      \begin{minipage}{0.6\textwidth}

\textbf{Parameters}: 
\begin{itemize}

\item Maximum weight allowed for the suitcase: 10 kg
\item Weight of the ski combination: 7 kg
\item Weight of the warm clothes: 4 kg
\item Weight of the hiking boots: 3 kg
\item Weight of the book on hiking: 1 kg
\item Weight of the umbrella: 2 kg
\end{itemize}

\textbf{Variables}:
\begin{itemize}
\item Binary variable x1: 1 if the ski combination is selected, 0 otherwise
\item Binary variable x2: 1 if the warm clothes are selected, 0 otherwise
\item Binary variable x3: 1 if the hiking boots are selected, 0 otherwise
\item Binary variable x4: 1 if the book on hiking is selected, 0 otherwise
\item Binary variable x5: 1 if the umbrella is selected, 0 otherwise
\end{itemize}

\textbf{Constraints}:
\begin{itemize}
\item The sum of the weights of the selected items cannot exceed the maximum weight allowed for the suitcase: x1*7 + x2*4 + x3*3 + x4*1 + x5*2 <= 10
\end{itemize}

\textbf{Objective function}:
\begin{itemize}
\item The objective is to maximize the importance of the selected item. We can introduce a weight for each item to represent its importance. Let w1 = 1 for the ski combination, w2 = 2 for the warm clothes, w3 = 3 for the hiking boots, w4 = 4 for the book on hiking, and w5 = 2 for the umbrella. Then, the objective is to maximize the weighted sum of the selected items: w1*x1 + w2*x2 + w3*x3 + w4*x4 + w5*x5.
\end{itemize}

      \end{minipage}
    };

    \node[fill=orange!20, draw, text width=0.7\textwidth, font=\footnotesize] (codebox) at (-5,-9.5) {
      \begin{verbatim}from cpmpy import *

# Parameters
max_weight = 10
weights = [7, 4, 3, 1, 2]
importance = [1, 2, 3, 4, 2]

# Variables
x = boolvar(shape=5)

# Constraints
model = Model(sum(x * weights) <= max_weight)

# Objective function
objective = sum(x * importance)
model.maximize(objective)

# Solve the problem
solution = model.solve()     \end{verbatim}
    };    
    
    \draw[->] (description) -- (formulation);
    \draw[->] (formulation) -- (codebox);

  \end{tikzpicture}
  \caption{Example with the 4th level of abstraction}
  \label{fig:ex2}
\end{figure}

\section{Summary and future work}
\label{sec:concl}

The Holy Grail of computer science has been defined as "the user states the problem, the computer solves it". With the help of the rich modeling languages developed recently, we are now closer to the stated goal. However, the gap between the natural description of a problem and its formal formulation as an optimization problem is still there.
The goal of this position paper is to outline our project of developing a framework able to take a description of a problem in natural language, model the problem, and finally turn it into an actual runnable constraint programming model. 
This framework would make use, among other things, of large language models such as GPT-3.5 or LLaMa in order to extract the features, identify the relation, formulate the problem, and then run it. 
The first tests show great potential for formalizing good constraint programming models, even when some parameters of the problem are only given in an abstract way. 

Future work should explore the usage of other LLMs, and also specialized methods for each subtask, like the ones in~\cite{dakle2023ner4opt,gangwar2022highlighting}. 
We can also exploit more optimization domain knowledge in prompt tuning, in-context learning, and fine-tuning. Fine-tuning LLMs for specific tasks has been shown to improve them~\cite{dodge2020fine}. Initial examples of fine-tuning LLMs with optimization-specific corpora show great potential~\cite{dakle2023ner4opt} and we would like to extend those results across the entire pipeline from text to final model. In addition, (soft) prompt tuning~\cite{lester2021power} can significantly improve the results, having been shown to outperform few-shot learning. Finally, different ways of interacting with the user should be examined, to correct system or user mistakes in a minimum number of interactions.

\bibliography{paper}

\end{document}